\documentclass{article}

\PassOptionsToPackage{numbers, compress}{natbib}

 \usepackage[preprint]{style_202x}

\usepackage[utf8]{inputenc} 
\usepackage[T1]{fontenc}    
\usepackage[pagebackref]{hyperref}       
\usepackage{url}            
\usepackage{booktabs}       
\usepackage{amsfonts}       
\usepackage{nicefrac}       
\usepackage{microtype}      
\usepackage{xcolor}         

\usepackage{graphicx}

\usepackage{amsmath}
\usepackage[capitalize]{cleveref}
\usepackage{multirow}
\usepackage{booktabs}
\usepackage{tabularx}
\usepackage{array}
\usepackage{makecell}
\usepackage{wrapfig}

\usepackage[table]{xcolor}

\definecolor{deepskyblue}{RGB}{54, 125, 189}
\definecolor{lightskyblue}{RGB}{58, 178, 198}
\hypersetup{colorlinks = true, linkcolor = deepskyblue,
            urlcolor  = lightskyblue,
            citecolor = deepskyblue,
            anchorcolor = deepskyblue}

\makeatletter
\newcommand{\ie}{\emph{i.e.}\@ifnextchar.{\!\@gobble}{}}
\newcommand{\eg}{\emph{e.g.}\@ifnextchar.{\!\@gobble}{}}
\newcommand{\etc}{etc\@ifnextchar.{}{.\@}}
\makeatother

\title{\underline{Con}trol Your Queries: Heterogeneous Query Interaction for Camera-Radar \underline{Fusion}}

\author{%
  Jialong Wu\textsuperscript{1,3} \quad
  Yihan Wang\textsuperscript{2,3} \quad
  Matthias Rottmann\textsuperscript{1} \\
  \textsuperscript{1}Osnabrück University \quad
  \textsuperscript{2}University of Wuppertal \quad
  \textsuperscript{3}Aptiv Services Deutschland GmbH
}

\begin{document}

\maketitle

\begin{abstract}

In autonomous driving, camera-radar fusion offers complementary sensing and low deployment cost.
Existing methods perform fusion through input mixing, feature map mixing, or query-based feature sampling.
We propose a new fusion paradigm, termed heterogeneous query interaction, and present \textbf{ConFusion}, a camera-radar 3D object detector. ConFusion combines image queries, radar queries, and learnable world queries distributed in 3D space to improve query initialization and object coverage.
To encourage cross-type interaction among heterogeneous queries, we introduce heterogeneous query mixing (\textbf{QMix}), which performs dedicated cross-type attention after feature sampling to consolidate complementary object evidence. We further propose interactive query swap sampling (\textbf{QSwap}), which improves feature sampling by allowing related queries to exchange informative feature tokens under attention and geometric constraints. Experiments on the nuScenes dataset show that ConFusion achieves state-of-the-art performance, reaching 59.1 mAP and 65.6 NDS on the validation set, and 61.6 mAP and 67.9 NDS on the test set.

\end{abstract}    
\section{Introduction} \label{sec:intro}

\begin{figure}[htbp] \centering
    \includegraphics[width=1.0\textwidth]{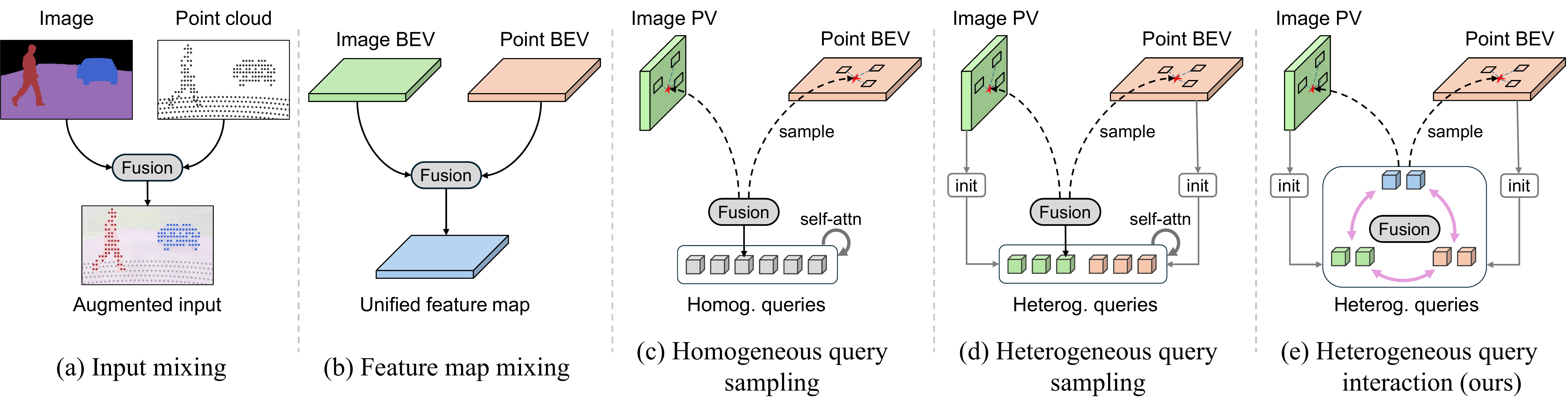}
    \caption{Evolution of camera-3D sensor fusion: (a) input mixing; (b) feature map mixing; (c,d) homogeneous and heterogeneous query sampling; and (e) our heterogeneous query interaction.} \label{fig:sec1_intro}
\end{figure}

Sensor fusion is a key direction in autonomous driving, as multiple sensors provide complementary cues and redundancy, improving robustness, accuracy, and reliability. A common setting combines cameras with 3D sensors such as LiDAR or radar \cite{FusionReviewOld}. Compared to LiDAR, radar is sparse and low resolution, yet it is more robust to adverse weather, provides velocity measurements, and is typically an order of magnitude cheaper \cite{mmwaveReview}. These properties make camera-radar fusion promising for large-scale deployment \cite{FusionReview1}. Despite these advantages, camera-radar fusion remains challenging, as the two modalities exhibit strong heterogeneity in both measurement space and representation. Therefore, developing fusion methods that can effectively associate multi-modal evidence is crucial. In this work, we focus on camera-radar fusion. We also discuss camera-3D sensor fusion more broadly, since camera-radar and camera-LiDAR fusion share common design patterns.

Figure \ref{fig:sec1_intro} summarizes how the definition of ``fusion'' has evolved in camera-3D sensor fusion for object detection. Early works interpret fusion as input-level mixing (Fig.\,\hyperref[fig:sec1_intro]{1a}), where point clouds are projected onto images and augmented with image semantics \cite{PointPainting, MVP}. Later methods fuse at the feature map level by mixing modality-specific features into a unified bird's-eye-view (BEV) representation (Fig.\,\hyperref[fig:sec1_intro]{1b}) \cite{BEVFusion, RCBEVDet}. In these feature map-level methods, fusion is performed when constructing the unified feature map, and predictions are produced from it using CNN- or transformer-based heads.

DETR-style detectors enable query-based fusion \cite{FUTR3D}, where object queries directly sample and aggregate object-related features from multi-modal feature maps (Fig.\,\hyperref[fig:sec1_intro]{1c}). In a typical decoder, queries are updated by self-attention and then attend to perspective-view (PV), point-cloud BEV, or image BEV features through cross-attention \cite{RaCFormer}.
Despite different query initialization strategies \cite{TransFusion, TransCAR}, these methods use a single query type, which we refer to as homogeneous query sampling.

In contrast, we refer to heterogeneous query sampling as the setting where multiple types of object queries are initialized from different sources (Fig.\,\hyperref[fig:sec1_intro]{1d}), \eg, combining image queries from an image detector with randomly scattered world queries \cite{SparseLIF}, or with LiDAR detector-derived queries \cite{MV2DFusion}. Such designs provide stronger initialization priors and improve object localization.

However, existing methods rely on conventional self-attention over the mixed query set, where cross-type information exchange is implicit \cite{SparseFusion,SpaRC,SparseFusion3D}. Given the strong heterogeneity between camera and radar modalities, we observe that self-attention tends to favor same-type queries. This same-type bias weakens cross-type evidence exchange and limits the benefits of heterogeneous queries.

In this paper, we push the definition of fusion further and propose a new fusion paradigm: fusion through heterogeneous query interaction (Fig.\,\hyperref[fig:sec1_intro]{1e}). Beyond sampling from multi-modal feature maps, we explicitly enforce cross-type interactions among heterogeneous queries. Focusing on camera-radar fusion, we employ three query sources, namely image queries, radar queries, and world queries scattered in 3D space, to improve initialization and recover missed objects. Multi-source initialization naturally yields redundant queries around the same object. We exploit this redundancy by encouraging information exchange among heterogeneous queries that are likely associated with the same object. Specifically, we introduce a heterogeneous query mixing (QMix) module. In each decoder layer, QMix is applied after queries have aggregated object-related features from multi-modal feature maps. QMix performs dedicated cross-type attention across the three query types, enabling queries to further consolidate complementary object evidence and better exploit heterogeneous initialization.

Besides embedding-level interaction, we further explore interaction at the geometric level and propose interactive query swap sampling (QSwap). During feature sampling, QSwap selects related queries based on attention affinities and allows them to exchange informative feature tokens, improving the quality of sampled object evidence. We impose both attention and geometric constraints to ensure reliable interactive sampling. Our model, ConFusion, achieves state-of-the-art performance on the nuScenes camera-radar detection benchmark \cite{nuScenes}, reaching 59.1 mAP and 65.6 NDS on the validation set, and 61.6 mAP and 67.9 NDS on the test set. Our main contributions are:
\begin{itemize}
    \item We propose a new fusion paradigm, fusion through heterogeneous query interaction, and introduce heterogeneous query mixing (QMix), which performs dedicated cross-type attention to consolidate complementary multi-source object evidence.
    \item We introduce interactive query swap sampling (QSwap), which improves feature sampling through constrained token sharing among related queries.
    \item ConFusion achieves state-of-the-art performance on nuScenes, improving the previous best by 1.8 mAP and 2.6 NDS on the validation set, and by 2.4 mAP and 2.0 NDS on the test set.
\end{itemize}

\section{Related Work}

\subsection{Transformer-based Object Detection}
DETR \cite{DETR} introduces the transformer paradigm to object detection by formulating a set of learnable object queries as latent prediction slots.
Deformable-DETR \cite{DeformableDETR} improves convergence and efficiency by predicting reference points from query embeddings and attending to a sparse set of feature samples around them.
Conditional-DETR \cite{ConditionalDETR} decouples each query into content and spatial components to improve localization.
DN-DETR \cite{DN-DETR} accelerates convergence through query denoising by injecting perturbed ground-truth boxes as additional queries. BEVFormer \cite{BEVFormer} proposes grid-based BEV queries to build a BEV representation from multi-view images.
DETR3D and PETR \cite{DETR3D,PETR} define 3D object queries to sample and aggregate object evidence from image PV feature maps.

\subsection{Camera and 3D Sensor Fusion for Object Detection}

\begin{table}[t]
\centering

\setlength{\tabcolsep}{5pt}
\setlength{\aboverulesep}{1.3pt}
\setlength{\belowrulesep}{1.3pt}

\caption{Camera-3D sensor fusion methods for object detection. C: camera; L: LiDAR; R: radar.
}
{\scriptsize 
\begin{tabularx}{0.94\textwidth}{@{}c c >{\centering\arraybackslash}X@{}}
\toprule
\textbf{Category} & \textbf{Input} & \textbf{Methods} \\
\midrule
Input mixing
& C+L &
EPNet \cite{EPNet}, PointPainting \cite{PointPainting}, PointAugmenting \cite{PointAugmenting}, FusionPainting \cite{FusionPainting}, MVP \cite{MVP}, PAI3D \cite{PAI3D} \\

\midrule
\multirow[c]{2}{*}[-2.3ex]{\makecell[c]{Feature map\\mixing}}
& C+L &
3D-CVF \cite{3D-CVF}, BEVFusion \cite{BEVFusion}, AutoAlignV2 \cite{AutoAlignv2}, DeepFusion \cite{DeepInteraction}, UVTR \cite{UVTR}, BEVFusion4D \cite{BEVFusion4D}, MSMDFusion \cite{MSMDFusion}, FusionFormer \cite{FusionFormer}, UniTR \cite{UniTR}, EA-LSS \cite{EA-LSS}, IS-Fusion \cite{IS-Fusion}, LAS \cite{LAS}, EVT \cite{EVT} \\
& C+R &
CenterFusion \cite{CenterFusion}, CramNet \cite{CramNet}, CRN \cite{CRN}, EchoFusion \cite{EchoFusion}, RCM-Fusion \cite{RCM-Fusion}, RCFusion \cite{RCFusion}, LXL \cite{LXL}, RCBEV \cite{RCBEV}, CRKD \cite{CRKD}, RCBEVDet \cite{RCBEVDet}, CRT-Fusion \cite{CRT-Fusion}, HGSFusion \cite{HGSFusion}, HyDRa \cite{Hydra} \\
\midrule

\multirow[c]{2}{*}{\makecell[c]{Homogeneous\\query sampling}}
& C+L &
TransFusion \cite{TransFusion}, DeepInteraction \cite{DeepInteraction}, FUTR3D \cite{FUTR3D}, CMT \cite{CMT}, ObjectFusion \cite{ObjectFusion}, DAL \cite{DAL} \\
& C+R &
TransCAR \cite{TransCAR}, CRAFT \cite{CRAFT}, DPFT \cite{DPFT}, RaCFormer \cite{RaCFormer}, CVFusion \cite{CVFusion} \\

\midrule

\multirow[c]{2}{*}{\makecell[c]{Heterogeneous\\query sampling}}
& C+L &
SparseFusion \cite{SparseFusion}, SparseLIF \cite{SparseLIF}, MV2DFusion \cite{MV2DFusion} \\
& C+R &
SparseFusion3D \cite{SparseFusion3D}, SpaRC \cite{SpaRC} \\

\midrule

\makecell[c]{Heterogeneous\\query interaction}
& C+R &
\makecell[c]{\textbf{ConFusion} (ours)} \\
\bottomrule
\end{tabularx}
}
\label{tab:sec2_fusion_methods}
\end{table}

Table~\ref{tab:sec2_fusion_methods} organizes prior camera-3D sensor fusion methods according to our taxonomy of ``fusion''.

\textbf{Input mixing.} Early fusion methods interpret fusion as input-level augmentation \cite{PAI3D,PointAugmenting,FusionPainting}.
PointPainting \cite{PointPainting} projects 3D points onto the image plane, appends image segmentation predictions to points, and applies a standard 3D detector on the augmented point cloud.
MVP \cite{MVP} increases point density on foreground objects by generating virtual points guided by 2D detectors.

\textbf{Feature map mixing.}
A second view of fusion is to construct a unified representation from modality-specific feature maps, using projection and concatenation \cite{3D-CVF, MSMDFusion, CramNet, CRKD, CRT-Fusion, HGSFusion}, cross-attention \cite{DeepFusion, CRN, EVT}, or BEV queries \cite{RCM-Fusion, Hydra}.
BEVFusion \cite{BEVFusion} fuses image BEV and point cloud BEV features through concatenation or weighted addition.
AutoAlignV2 \cite{AutoAlignv2} aligns modality-specific BEV features via deformable cross-attention, while RCBEVDet \cite{RCBEVDet} adopts bidirectional cross-attention.
FusionFormer \cite{FusionFormer} improves the unified BEV feature by leveraging BEV queries.
EchoFusion \cite{EchoFusion} applies BEV queries to fuse radar raw data with images.
UniTR \cite{UniTR} directly constructs a unified token sequence with a modality-agnostic encoder.
IS-Fusion \cite{IS-Fusion} couples object proposals with unified BEV features through bidirectional cross-attention, enhancing the unified BEV features.

\textbf{Homogeneous query sampling.}
DETR-style detectors enable query-based sampling, where a single type of object query performs fusion by directly sampling and aggregating object-related features from multi-modal feature maps \cite{DeepInteraction, DAL, TransCAR, CRAFT}. FUTR3D \cite{FUTR3D} first applies self-attention among queries, then uses deformable cross-attention to sample feature tokens and update the query embeddings.
CMT \cite{CMT} allows queries to attend to dense multi-modal tokens augmented with geometry-aware positional encoding.
TransFusion \cite{TransFusion} and ObjectFusion \cite{ObjectFusion} adopt heatmap-based query initialization to improve proposal quality.
DPFT \cite{DPFT} performs feature aggregation over multi-view radar feature maps.
RaCFormer \cite{RaCFormer} designs a circular query initialization scheme to match ray-based sensing geometry.
CVFusion \cite{CVFusion} first extracts RoI proposals from unified BEV features and then uses them as queries to aggregate image, point-cloud, and unified BEV features.

\textbf{Heterogeneous query sampling.}
Recent methods have begun to use multiple query types with different initializations, providing stronger priors for object queries.
SparseFusion \cite{SparseFusion} performs fusion in two stages. It first applies feature map mixing between the image PV and LiDAR BEV branches without constructing a unified feature map. It then initializes image and LiDAR queries, separately samples modality-specific features, and applies self-attention over the heterogeneous query set.
SparseLIF \cite{SparseLIF} combines image queries with randomly scattered world queries and injects sensor uncertainty during feature aggregation.
MV2DFusion \cite{MV2DFusion} employs image and LiDAR queries and explicitly models depth uncertainty for image queries.
SpaRC \cite{SpaRC} performs local self-attention over image and world queries after aggregating point-wise radar features and image PV features.

Although multi-source initialization provides stronger object coverage, existing methods still rely on standard self-attention, leaving cross-type information exchange implicit.
In this paper, we propose a new fusion paradigm that treats fusion as explicit interaction among heterogeneous queries.
\section{ConFusion Architecture} \label{sec:method}

\begin{figure}[htbp] \centering
    \includegraphics[width=1.0\textwidth]{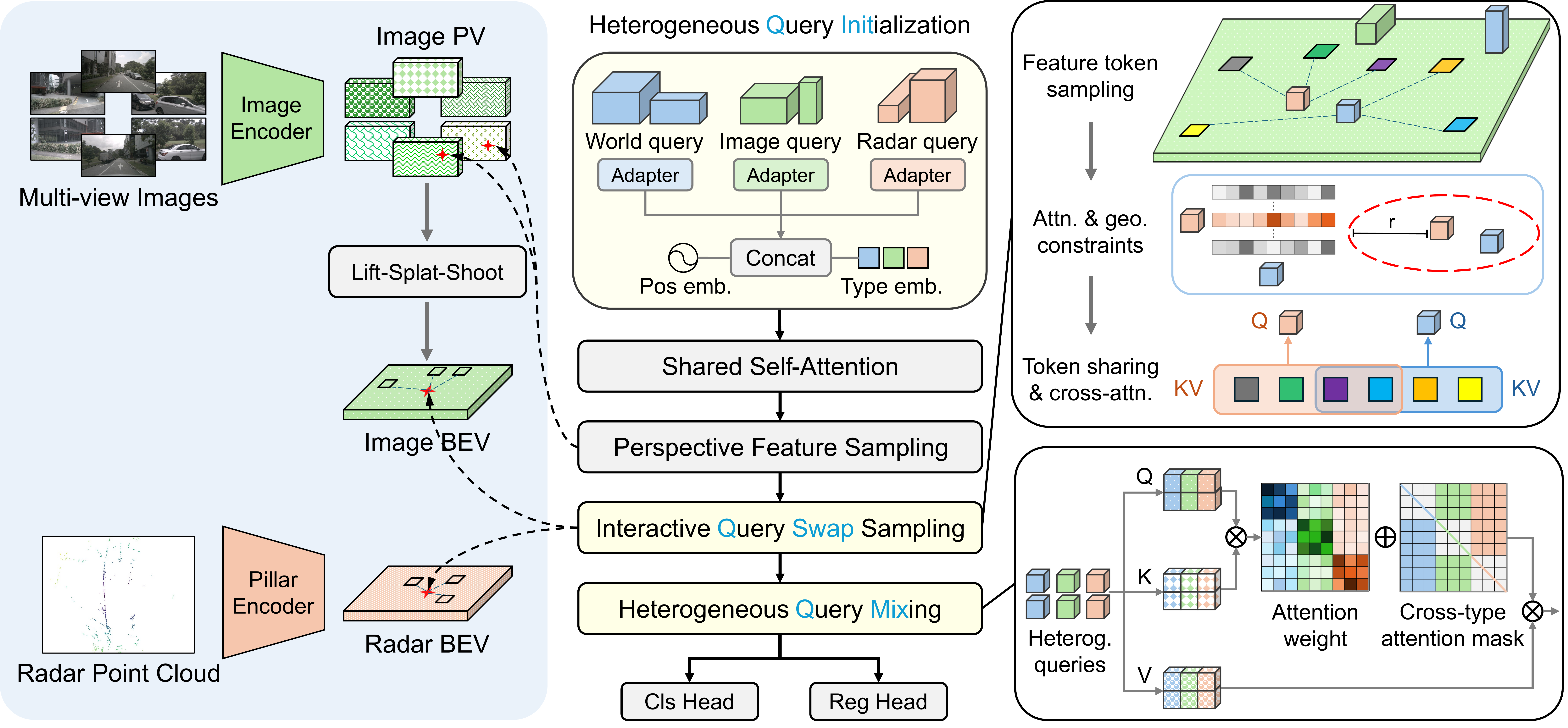}
    \caption{Overview of ConFusion. ConFusion first extracts multi-modal feature maps from camera and radar inputs, and then initializes heterogeneous image, radar, and world queries. The queries are decoded by an $L$-layer transformer decoder, where QSwap enables constrained token sharing during feature sampling, while QMix performs dedicated cross-type attention after feature aggregation.} \label{fig:sec3_archi}
\end{figure}

In this section, we present ConFusion, a camera-radar 3D detector based on heterogeneous query interaction. The overall architecture is shown in \cref{fig:sec3_archi}. ConFusion first extracts multi-modal feature maps from multi-view images and radar point clouds (\cref{sec:3.1}), and then performs heterogeneous query initialization with image, radar, and world queries (\cref{sec:3.2}). These queries are fed into an $L$-layer transformer decoder, where fusion is achieved through two complementary query interaction modules: heterogeneous query mixing (QMix), which consolidates object evidence among multi-source queries after feature aggregation (\cref{sec:3.3}), and interactive query swap sampling (QSwap), which enables queries to exchange informative feature tokens during feature sampling (\cref{sec:3.4}). At each layer, detection heads generate intermediate class and box predictions from updated queries.

\subsection{Backbone Networks}  \label{sec:3.1}
    
ConFusion takes multi-view images and radar point clouds as input. The backbone networks produce three feature maps: image PV features $\mathbf{F}_{\mathrm{img}}^{\mathrm{pv}}$, image BEV features $\mathbf{F}_{\mathrm{img}}^{\mathrm{bev}}$, and radar BEV features $\mathbf{F}_{\mathrm{rad}}^{\mathrm{bev}}$. In the camera branch, a CNN backbone with an FPN extracts multi-view image PV features, which are then lifted into BEV through pixel-wise depth prediction using the lift-splat-shoot pipeline \cite{LSS}. In the radar branch, a PointPillars-style encoder \cite{PointPillars} generates radar BEV features. These feature maps serve as scene representations for subsequent query initialization and transformer decoding.

\subsection{Heterogeneous Query Initialization}  \label{sec:3.2}

Object queries can be viewed as modeling of target objects in terms of semantics and location \cite{ConditionalDETR}. In multi-modal fusion, we initialize queries from multiple sources with different priors, so that each target object can be modeled from complementary cues. This motivates a new fusion paradigm that treats fusion as explicit interaction among heterogeneous yet complementary object queries. Specifically, we define three query types: image queries, radar queries, and world queries. We use $\mathbf{Q}_{*}^{\mathrm{emb}} \in \mathbb{R}^{N_*^Q \times d}$ to denote query embeddings and $\mathbf{Q}_{*}^{\mathrm{pos}} \in \mathbb{R}^{N_*^Q \times 3}$ to denote 3D query positions.

For image queries, we exploit the rich semantic cues of cameras. Image queries are initialized from 2D proposals predicted on the image PV features by a lightweight 2D CNN-based detection head. The corresponding PV feature vectors serve as the initial query embeddings $\mathbf{Q}_{\mathrm{img}}^{\mathrm{emb}}$, while the 2D proposal centers are lifted into 3D through depth prediction to obtain the initial positions $\mathbf{Q}_{\mathrm{img}}^{\mathrm{pos}}$.

For radar queries, we exploit the geometric and velocity cues of radars. Radar queries are initialized from top-scoring cells predicted by a lightweight heatmap head on the radar BEV features. The associated BEV feature vectors are used as $\mathbf{Q}_{\mathrm{rad}}^{\mathrm{emb}}$, and the cell centers define $\mathbf{Q}_{\mathrm{rad}}^{\mathrm{pos}}$. Since radar is less reliable for precise shape estimation, we use the heatmap head only to predict classification scores, and leave precise box regression to the subsequent fusion-based transformer decoder.

\begin{figure}[htbp] \centering
    \includegraphics[width=0.80\textwidth]{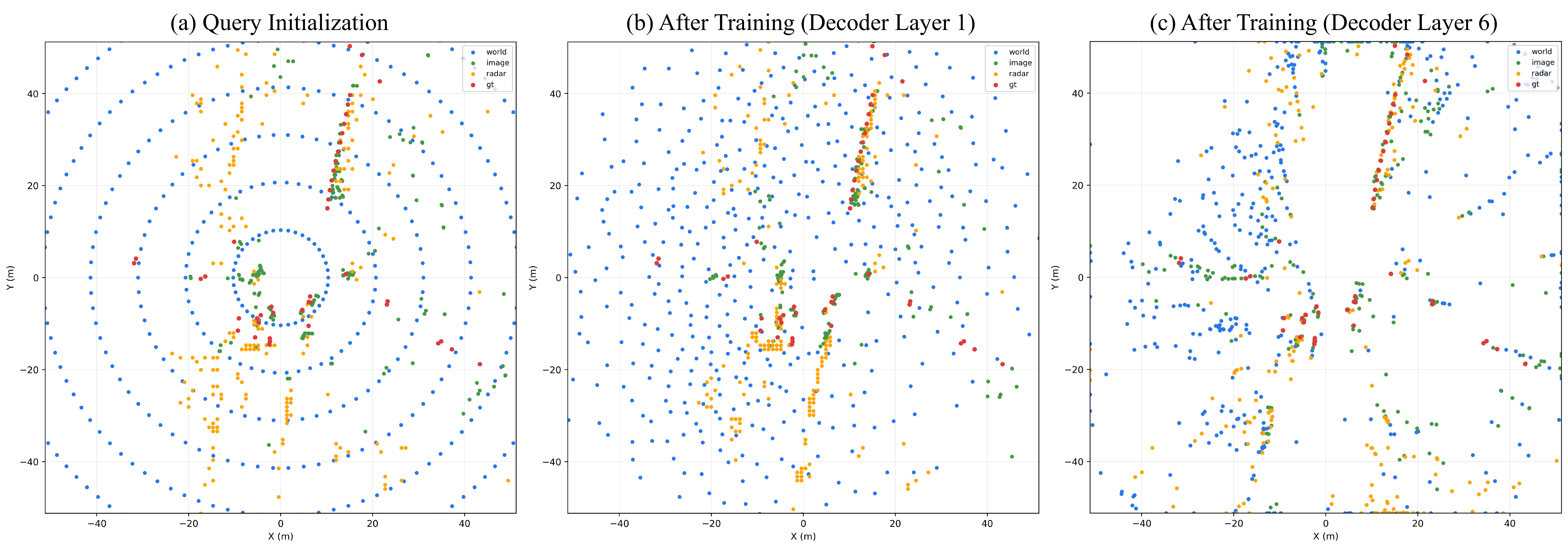}
    \caption{Heterogeneous query initialization and post-training distributions across decoder layers.} \label{fig:sec3_query_init}
\end{figure}

For world queries, we additionally introduce a set of learnable queries to improve object coverage, since single-modality pre-detectors inevitably miss some objects. Following \cite{RaCFormer, RayFormer}, world queries are initialized on concentric circles with increasing density at larger ranges. Their positions $\mathbf{Q}_{\mathrm{w}}^{\mathrm{pos}}$ are optimized during training, and their embeddings $\mathbf{Q}_{\mathrm{w}}^{\mathrm{emb}}$ are learnable. \Cref{fig:sec3_query_init} visualizes the distributions of the three query types. Image and radar queries are initialized around potential objects, while world queries become more broadly distributed across the scene after training.

The heterogeneous queries are then fed into the decoder. In each decoder layer, we first apply a type-specific adapter to each query group, encouraging different query types to preserve their distinct roles. We further add learnable type embeddings to explicitly encode query identity before the subsequent shared self-attention (\cref{fig:sec3_archi}). The queries then perform deformable feature sampling over image PV, image BEV, and radar BEV feature maps, followed by cross-attention for feature aggregation. We will later revisit the sampling stage in \cref{sec:3.4} when introducing QSwap. Before that, we first present QMix, which focuses on heterogeneous query interaction after feature aggregation.

\subsection{Heterogeneous Query Mixing (QMix)}  \label{sec:3.3}

Multi-source query initialization naturally yields redundant queries around the same object. Moreover, as the decoder iteratively refines query embeddings and positions across $L$ layers, queries, especially world queries, are gradually attracted toward nearby object locations, as illustrated in Fig.\,\hyperref[fig:sec3_query_init]{3b-c}. This creates opportunities for heterogeneous queries with different priors that correspond to the same object to exchange information. We therefore interpret post-aggregation interaction among heterogeneous queries as fusion at the object-evidence level.

\begin{figure}[htbp] \centering
    \includegraphics[width=0.70\textwidth]{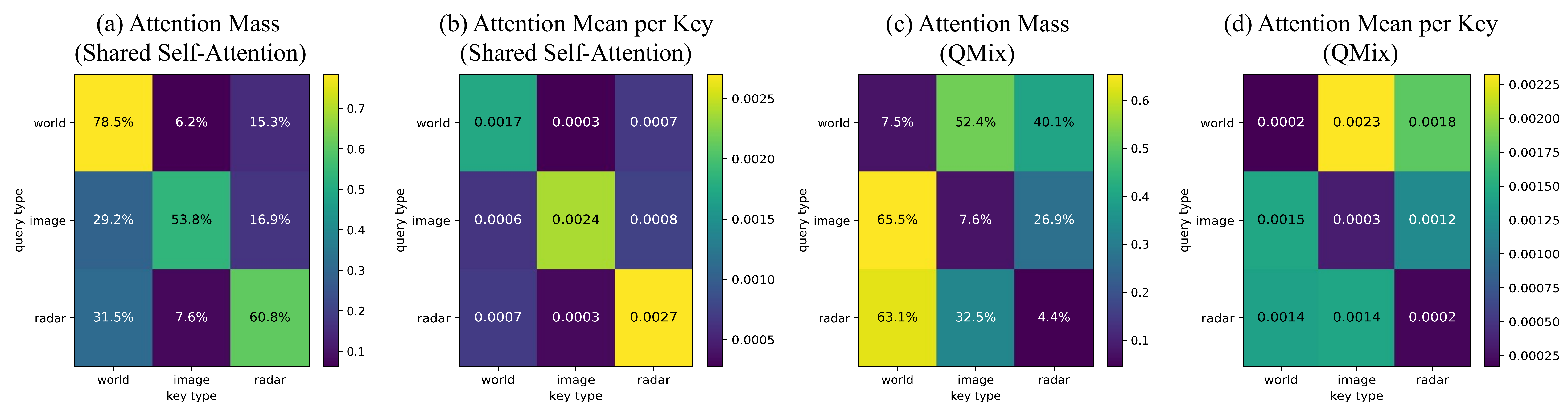}
    \caption{Attention statistics of shared self-attention and QMix across query types.} \label{fig:sec3_qmix}
\end{figure}

Existing heterogeneous-query fusion methods typically apply standard self-attention over the mixed query set, either before feature sampling \cite{MV2DFusion, SparseLIF} or after feature aggregation \cite{SparseFusion, SpaRC}. In our framework, although type-specific adapters and type embeddings help distinguish query types, shared self-attention still does not provide sufficient cross-type information exchange. We observe a pronounced same-type bias in the shared self-attention: image, radar, and world queries preferentially attend to queries of the same type. As shown in \cref{fig:sec3_qmix}, we report both the type-to-type attention mass and the mean per-key attention weight. Since $N_{\mathrm{w}}^{Q}>N_{\mathrm{img}}^{Q}=N_{\mathrm{rad}}^{Q}$, the former is affected by query-count imbalance, while the latter is averaged per key. This observation motivates our heterogeneous query mixing (QMix) module, which explicitly promotes heterogeneous query interaction.

In each decoder layer, after queries aggregate object evidence from multi-modal feature maps, we insert a dedicated cross-type attention layer. A straightforward design is to perform separate pairwise cross-attention among the three query types, but this is computationally expensive and empirically unstable. QMix instead uses masked multi-head attention to promote cross-type interaction (\cref{fig:sec3_archi}).
Specifically, we block attention between distinct queries of the same type, forcing each query to interact only with heterogeneous queries of the other two types. We keep the diagonal entries open, so that a query can still attend to itself when no reliable heterogeneous partner exists.

Formally, let $\mathbf{Q}=\mathrm{Concat}(\mathbf{Q}_{\mathrm{img}}^{\mathrm{emb}},\mathbf{Q}_{\mathrm{rad}}^{\mathrm{emb}},\mathbf{Q}_{\mathrm{w}}^{\mathrm{emb}})$ denote the concatenated query embeddings of all three types at the current decoder layer, and let $c_i \in \{\mathrm{img}, \mathrm{rad}, \mathrm{w}\}$ denote the type of the $i$-th query. We define the cross-type attention mask $\mathbf{M} \in \mathbb{R}^{N^Q \times N^Q}$ as
\begin{equation}
M_{ij} =
\begin{cases}
0, & i=j \ \text{or}\ c_i \neq c_j,\\
-\infty, & i\neq j \ \text{and}\ c_i=c_j.
\end{cases}
\end{equation}
QMix then applies masked multi-head attention as
\begin{equation}
\mathbf{Q}' = \mathrm{MHA}(\mathbf{Q}, \mathbf{Q}, \mathbf{Q}; \mathbf{M}),
\end{equation}
where $\mathrm{MHA}$ denotes multi-head attention and $\mathbf{M}$ is the cross-type attention mask. The updated queries are then fed into an MLP followed by the detection head of the current decoder layer.

By enforcing cross-type interaction at the embedding level, QMix enables heterogeneous queries with different priors to fuse object evidence gathered from multi-modal feature maps. As shown in \cref{fig:sec3_qmix}, QMix substantially increases cross-type interaction in comparison to standard self-attention.

\subsection{Interactive Query Swap Sampling (QSwap)}  \label{sec:3.4}

While QMix performs interaction at the query embedding level after feature aggregation, we further explore query interaction at the geometric level during feature sampling. To this end, we propose interactive query swap sampling (QSwap), which improves the quality of sampled object evidence by allowing each query to incorporate informative tokens from nearby related queries.

Specifically, as shown in \cref{fig:sec3_archi}, each query first predicts $K_{\mathrm{base}}$ base sampling offsets and their scores around its current position, as in deformable attention \cite{DeformableDETR}. QSwap then selects interaction neighbors using both attention affinities and geometric constraints. It augments each query's sampling set with high-score tokens shared from these neighbors for subsequent cross-attention feature aggregation.

To identify interaction neighbors, we use the attention weights from the shared self-attention as a query affinity prior. For each query, we first select the top-$N$ neighboring queries with the highest attention weights. Since attention alone may connect distant queries, we further impose a geometric constraint and only keep neighbors whose BEV center distance is within a query-adaptive radius $r_i = \alpha \cdot \sqrt{w_i^2 + l_i^2},$
where $w_i$ and $l_i$ denote the intermediate width and length predicted from query $i$. Since multi-scale perspective-view sampling \cite{SparseBEV} is more sensitive to geometric perturbations, we apply QSwap only to image and radar BEV feature sampling, indicated by dashed arrows in \cref{fig:sec3_archi}.

After selecting valid neighbors, each query augments its own sampling set with extra tokens shared from its neighbors. Specifically, let $s_{j,k}$ denote the predicted score assigned by neighbor query $j$ to its $k$-th base sampling point, and let $a_{ij}$ denote the affinity prior between query $i$ and query $j$ derived from shared self-attention. We rank neighbor-sourced sampling points by
\begin{equation}
\tilde{s}_{ijk} = s_{j,k} + \lambda \log a_{ij},
\end{equation}
where $\lambda$ controls the strength of the attention prior. To avoid domination by a single neighbor, we allow each neighbor to contribute at most $K_{\mathrm{per}}$ extra points, while each query receives at most $K_{\mathrm{extra}}$ shared points in total. These points are then integrated into the original sampling set, either by appending them or by replacing low-score base sampling points. Their scores are jointly normalized with the $K_{\mathrm{base}}$ original sampling scores before cross-attention feature aggregation.

In this way, QSwap improves feature sampling by allowing related queries to share high-score evidence tokens under both attention and geometric constraints. As a result, each query can collect more reliable object evidence, making QSwap complementary to QMix, which operates after feature aggregation to consolidate object evidence among heterogeneous queries.

\section{Experiments} \label{sec:exp}
\subsection{Dataset and Metrics} \label{sec:4.1}

We evaluate ConFusion on nuScenes \cite{nuScenes}, which contains 1,000 scenes and about 40k keyframes, with a 700/150/150 split for training, validation, and testing. The dataset provides 6 surround-view cameras and 5 FMCW radars, with 3D annotations at 2 Hz for 10 object classes. Following the official evaluation, we report mean average precision (mAP) and nuScenes detection score (NDS) as the main metrics. NDS is a weighted combination of mAP, mean average translation error (mATE), scale error (mASE), orientation error (mAOE), velocity error (mAVE), and attribute error (mAAE).

\subsection{Implementation Details} \label{sec:4.2}

We follow RaCFormer \cite{RaCFormer} for the training settings. We use ResNet-50 and ResNet-101 \cite{ResNet} backbones on the nuScenes validation set, and VoVNet (V2-99) \cite{VOV99} for test-set submission. We use one keyframe with 7 historical sweeps as input. The voxel size for image and radar BEV is [0.8, 0.8] m. We train with AdamW using a learning rate of $4\times10^{-4}$ and a batch size of 8. The ResNet-50 model is trained for 36 epochs, while the other backbones for 24 epochs. All models are trained on 4 A100 GPUs. The training loss includes decoder classification and box regression losses, together with the 2D pre-detector loss and radar heatmap loss. Full loss definitions are provided in Appendix.

We use 450 world, 225 image, and 225 radar queries, with the same total count of 900 queries as \cite{RaCFormer, SpaRC}. Image queries are generated by a YOLOX head \cite{Far3D}, with 50 proposals per view before selecting the top 225. The YOLOX head and radar heatmap head are pre-trained for 6 epochs only. The transformer decoder has 6 layers with shared weights for efficiency. For QSwap, we set the radius factor $\alpha=1.5$, attention prior strength $\lambda=1.0$, and base BEV sampling points $K_{\mathrm{base}}=20$. We set $K_{\mathrm{per}}=2$ and $K_{\mathrm{extra}}=4$. For ResNet-50, QSwap appends these shared points. For ResNet-101 and VoVNet, QSwap replaces low-score base sampling points to keep the final sampled points at $K_{\mathrm{base}}$.

\begin{table}[htbp]
\caption{Comparison with existing methods on the nuScenes \texttt{val} set. C: camera, R: radar. 
}
\centering
\setlength\tabcolsep{6pt}
\resizebox{1.0\textwidth}{!}{
\footnotesize
\begin{tabular}{l|cccc|cc|ccccc}
\toprule
\toprule
Methods & Input &Image Size & Backbone & Epochs & \textbf{mAP}$\uparrow$ & \textbf{NDS}$\uparrow$ & mATE$\downarrow$  & mASE$\downarrow$   & mAOE$\downarrow$  & mAVE$\downarrow$  &  mAAE$\downarrow$  \\
\midrule
RayDN \cite{RayDN} & C  &256$\times$704    & ResNet50 & 60 & 46.9    & 56.3    & 0.579 & 0.264 & 0.433 & 0.265 & 0.187     \\
Sparse4Dv3 \cite{Sparse4Dv3} & C  &256$\times$704    & ResNet50 & 100 & 46.9    & 56.1    & 0.553 & 0.274 & 0.476 & 0.227 & 0.200     \\
\midrule
HVDetFusion \cite{HVDetFusion} & C+R  & 256$\times$704    & ResNet50 &  24  & 45.1    & 55.7    & 0.557 & 0.527 & 0.270 & 0.473 & 0.212     \\
RCBEVDet \cite{RCBEVDet} & C+R  & 256$\times$704    & ResNet50 &  12  & 45.3  & 56.8   & 0.486 & 0.285 & 0.404 & 0.220 & 0.192     \\
CRN \cite{CRN} & C+R  & 256$\times$704    & ResNet50 &  24 & 49.0  & 56.0   & 0.487 & 0.277 & 0.542 & 0.344 & 0.197     \\
HyDRa~\cite{Hydra} & C+R  & 256$\times$704    & ResNet50 &  20 & 49.4  & 58.5   & 0.463 & 0.268 & 0.478 & 0.227 & 0.182     \\
RQR3D \cite{RQR3D} & C+R &256$\times$704   & ResNet50 & 20 &  50.7  &   59.2  & 0.437  &  0.289 & 0.435  &  0.232 & 0.193 \\
CRT-Fusion \cite{CRT-Fusion} & C+R &256$\times$704   & ResNet50 & 24 &  50.0  &   57.2  & 0.499  &  0.277 & 0.531  &  0.261 & 0.192 \\
SpaRC \cite{SpaRC} & C+R &256$\times$704   & ResNet50 & 60 &  54.5  &   62.0  & 0.496  &  0.269 & 0.403  &  0.177 & 0.181 \\
RaCFormer \cite{RaCFormer} & C+R &256$\times$704   & ResNet50 & 36 &  54.1 &   61.3  & 0.478  &  0.261 & 0.449  &  0.208 & 0.180 \\
\rowcolor[gray]{.92}\textbf{ConFusion} (Ours) & C+R &256$\times$704   & ResNet50 & 36 &  \textbf{55.9}  &   \textbf{63.1}  & 0.463  &  0.262 & 0.372  &  0.203 & 0.180 \\

\midrule
\midrule

RayDN \cite{RayDN} & C  &512$\times$1408    & ResNet101 & 60 & 51.8 & 60.4 & 0.541 & 0.260 & 0.315 & 0.236 & 0.200     \\  
Sparse4Dv3 \cite{Sparse4Dv3} & C  &512$\times$1408    & ResNet101 & 100 & 53.7    & 62.3    & 0.511 & 0.255 & 0.306 & 0.194 & 0.192     \\

\midrule

CRN \cite{CRN} & C+R  &512$\times$1408    & ResNet101 &  24 & 52.5 & 59.2 & 0.460 & 0.273 & 0.443 & 0.352 & 0.180     \\
HyDRa \cite{Hydra} & C+R  & 512$\times$1408    & ResNet101 & 20  & 53.6 & 61.7 & 0.416 & 0.264 & 0.407 & 0.231 & 0.186     \\
RICCARDO \cite{RICCARDO} & C+R & 512$\times$1408   & ResNet101 & / &  54.4  &   62.2  & 0.481  &  0.266 & 0.325  &  0.237 & 0.189 \\
RQR3D \cite{RQR3D} & C+R & 512$\times$1408   & ResNet101 & 20 &  54.7  &   62.2  & 0.417  &  0.281 & 0.381  &  0.228 & 0.188 \\
CRT-Fusion \cite{CRT-Fusion} & C+R & 512$\times$1408   & ResNet101 & 24 &  55.4  &   62.1  & 0.425  &  0.264 & 0.433  &  0.237 & 0.193 \\
SpaRC \cite{SpaRC} & C+R & 512$\times$1408   & ResNet101 & 60 &  57.1  &   64.4  & 0.484  &  0.264 & 0.308  &  0.175 & 0.178 \\
RaCFormer \cite{RaCFormer} & C+R  &512$\times$1408   & ResNet101 & 24 & 57.3 & 63.0 & 0.476 & 0.261 & 0.428 & 0.213 & 0.183   \\  
\rowcolor[gray]{.92}\textbf{ConFusion} (Ours) & C+R  &512$\times$1408   & ResNet101 & 24 & \textbf{59.1} & \textbf{65.6} & 0.448 & 0.255 & 0.310 & 0.192 & 0.188   \\  
\bottomrule
\bottomrule
\end{tabular}
}
\label{tab:sec4_valset}
\end{table}

\subsection{Main Results}  \label{sec:4.3}

\cref{tab:sec4_valset} compares ConFusion with previous methods on the nuScenes validation set. Under the ResNet-50, 256$\times$704 setting, ConFusion achieves 55.9\% mAP and 63.1\% NDS. Under the same training setup, it surpasses RaCFormer \cite{RaCFormer}, a homogeneous-query baseline, by 1.8 mAP and 1.8 NDS. ConFusion also outperforms SpaRC \cite{SpaRC}, a heterogeneous-query sampling method that still relies on implicit self-attention, by 1.4 mAP and 1.1 NDS, while requiring fewer training epochs.
Under the ResNet-101, 512$\times$1408 setting, ConFusion achieves 59.1\% mAP and 65.6\% NDS. It outperforms RaCFormer by 1.8 mAP and 2.6 NDS, establishing a new state of the art. Compared with camera-only Sparse4Dv3 \cite{Sparse4Dv3}, ConFusion demonstrates the benefit of sensor fusion.

\begin{table}[htbp]
\caption{Comparison with existing methods on the nuScenes \texttt{test} set. C: camera; R: radar. 
}
\centering
\setlength\tabcolsep{6pt}
\resizebox{1.0\textwidth}{!}{
\footnotesize
\begin{tabular}{l|cccc|cc|ccccc}
\toprule
\toprule
Methods & Input &Image Size & Backbone & Epochs & \textbf{mAP}$\uparrow$ & \textbf{NDS}$\uparrow$ & mATE$\downarrow$  & mASE$\downarrow$   & mAOE$\downarrow$  & mAVE$\downarrow$  &  mAAE$\downarrow$  \\
\midrule

RCBEVDet \cite{RCBEVDet} & C+R  & 640$\times$1600    & V2-99 &  12  & 55.0  & 63.9   & 0.390 & 0.234 & 0.362 & 0.259 & 0.113     \\
CRN \cite{CRN} & C+R  &  640$\times$1600    & ConvNeXt-B & 24  & 57.5  & 62.4   & 0.416 & 0.264 & 0.456 & 0.365 & 0.130    \\
HyDRa \cite{Hydra} & C+R  &  640$\times$1600   &  V2-99 & 20  & 57.4  & 64.2   & 0.398 & 0.251 & 0.423 & 0.249 & 0.122     \\
CRT-Fusion \cite{CRT-Fusion} & C+R  &  640$\times$1600   & ConvNeXt-B & 24 & 58.3 & 64.9 & 0.365 & 0.261 & 0.405 & 0.262 & 0.132  \\
SpaRC \cite{SpaRC} & C+R  &  640$\times$1600   & V2-99 & 60 & 60.0 & 67.1 &   &   &N/A   &   &       \\
RaCFormer \cite{RaCFormer} & C+R  &  640$\times$1600   & V2-99 & 24 & 59.2 & 65.9 & 0.407 & 0.244 & 0.345 & 0.238 & 0.132     \\
\rowcolor[gray]{.92}\textbf{ConFusion} (Ours) & C+R  &  640$\times$1600   & V2-99 & 24 & \textbf{61.6} & \textbf{67.9} & 0.382 & 0.236 & 0.348 & 0.201 & 0.123   \\
\bottomrule
\bottomrule
\end{tabular}
}
\label{tab:sec4_testset}
\end{table}

\begin{figure}[htbp] \centering
    \includegraphics[width=1.0\textwidth]{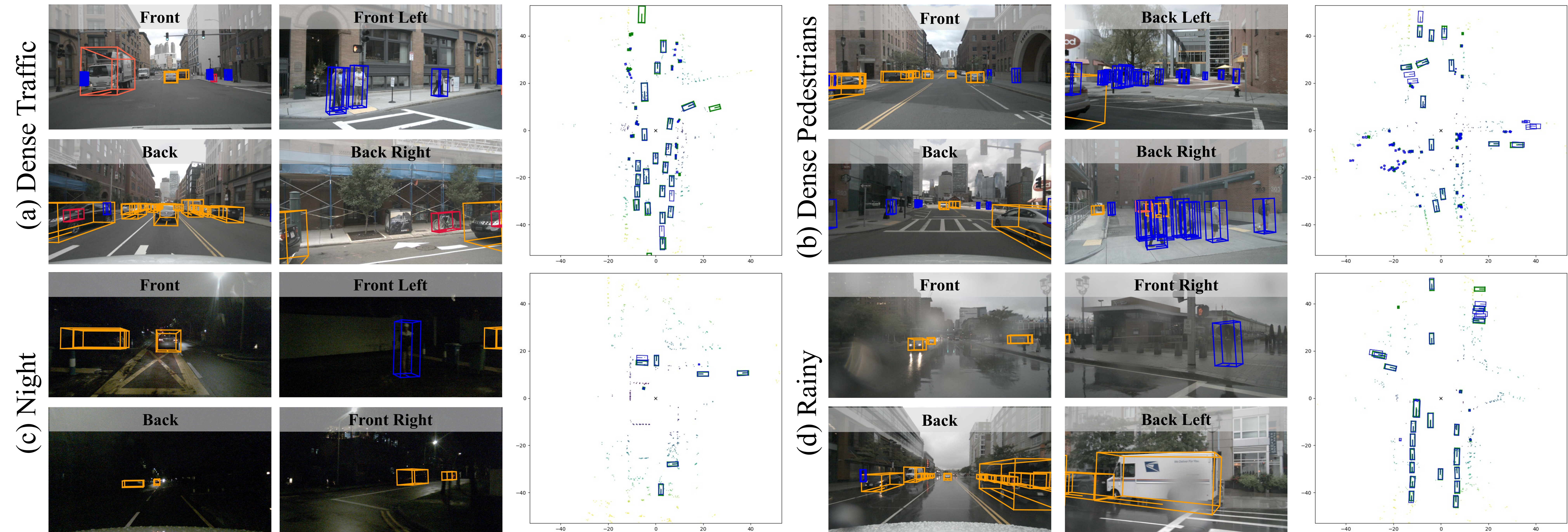}
    \caption{Qualitative results on the nuScenes \texttt{val} set under challenging scenarios. Left: projected 3D detections on multi-view images. Right: BEV radar points, predictions in blue, ground truth in green.} \label{fig:sec4_viz_result}
\end{figure}

\Cref{tab:sec4_testset} compares ConFusion with previous methods on the nuScenes test set. Under the same training setup as RaCFormer \cite{RaCFormer}, using a VoVNet backbone pre-trained on DD3D \cite{DD3D} and an input resolution of 640$\times$1600, ConFusion achieves 61.6\% mAP and 67.9\% NDS. It surpasses RaCFormer by 2.4 mAP and 2.0 NDS. Compared with SpaRC \cite{SpaRC}, ConFusion also improves mAP by 1.6 and NDS by 0.8, while using substantially fewer training epochs (24 vs.\ 60). \Cref{fig:sec4_viz_result} shows qualitative results in dense-object scenes and adverse conditions. Further discussion of failure cases and runtime analysis is provided in Appendix.

\subsection{Ablation Studies}

We conduct ablation studies under the same setup as the main results, using ResNet-50 and training for 36 epochs. \textit{QInit} denotes heterogeneous query initialization only, without QMix or QSwap. \textit{QMix} adds the QMix module on top of QInit, and \textit{QSwap} further adds QSwap on top of QMix.

\begin{table*}[htbp]
\centering
\begin{minipage}[t]{0.562\textwidth}
\centering
\caption{Ablation study of key components.}
\resizebox{\textwidth}{!}{
\begin{tabular}{l|c|ccccc}
\toprule
Methods & $N^Q$ & \textbf{mAP}$\uparrow$ & \textbf{NDS}$\uparrow$ & mATE$\downarrow$ & mASE$\downarrow$ & mAOE$\downarrow$ \\
\midrule
RaCFormer \cite{RaCFormer} & 900 & 54.14 & 61.33 & 0.478 & 0.261 & 0.449 \\
QInit & 900 & 54.88 & 61.66 & 0.473 & 0.265 & 0.447 \\
+QMix & 900 & 55.51 & 62.65 & 0.457 & 0.259 & 0.405 \\
\rowcolor[gray]{.92}+QMix+QSwap & 900 & \textbf{55.89} & \textbf{63.14} & 0.463 & 0.262 & 0.372 \\
\bottomrule
\end{tabular}
}
\label{tab:sec4_ablation_main}
\end{minipage}
\hfill
\begin{minipage}[t]{0.425\textwidth}
\centering
\caption{Ablation of query composition.}
\resizebox{\textwidth}{!}{
\begin{tabular}{l|c|ccc|cc}
\toprule
Setting & $N^Q$ & $N_{\mathrm{w}}^{Q}$ & $N_{\mathrm{img}}^{Q}$ & $N_{\mathrm{rad}}^{Q}$ & \textbf{mAP}$\uparrow$ & \textbf{NDS}$\uparrow$ \\
\midrule
QMix & 750 & 450 & 150 & 150 & 55.19 & 62.37 \\
QMix & 900 & 300 & 300 & 300 & 55.27 & 62.36 \\
\rowcolor[gray]{.92}QMix & 900 & 450 & 225 & 225 & 55.51 & 62.65 \\
QMix & 1050 & 600 & 225 & 225 & \textbf{55.87} & \textbf{62.91} \\
\bottomrule
\end{tabular}
}
\label{tab:sec4_ablation_qinit}
\end{minipage}

\end{table*}

\textbf{Performance Breakdown.} \cref{tab:sec4_ablation_main} shows the contribution of each component. Replacing the 900 world queries in RaCFormer \cite{RaCFormer} with three heterogeneous query types already achieves 54.88\% mAP and 61.66\% NDS. This suggests that heterogeneous query initialization provides better query position and embedding priors. Adding QMix improves the performance to 55.51\% mAP and 62.65\% NDS, corresponding to gains of 0.63 mAP and 0.99 NDS over QInit. In particular, mATE drops from 0.473 to 0.457 and mAOE from 0.447 to 0.405, indicating that explicit cross-type interaction not only improves detection accuracy but also enhances box localization and orientation estimation. Adding QSwap further improves the performance to 55.89\% mAP and 63.14\% NDS, bringing another 0.38 mAP and 0.49 NDS gains. Meanwhile, mAOE decreases from 0.405 to 0.372, showing that QSwap improves the quality of sampled object cues and provides complementary benefits to QMix.

\textbf{QInit Analysis.} \Cref{tab:sec4_ablation_qinit} studies different query compositions with QMix enabled. Compared with QInit and RaCFormer \cite{RaCFormer}, the setting with only 750 queries already improves performance, showing the benefit of QMix. Among the 900-query settings, 450/225/225 performs best, suggesting that sufficient world queries are important for covering missed objects. Increasing the number of queries to 1050 further improves performance, suggesting favorable scaling behavior of ConFusion.

\textbf{QMix Analysis.}
\cref{tab:sec4_ablation_qmix} compares different QMix insertion strategies. All settings include QInit and shared self-attention by default. Adding an extra self-attention layer after feature aggregation (+Post-self) does not improve performance over QInit. Adding cross-type attention after this extra self-attention further degrades mATE and mAOE, suggesting that repeated query interaction after aggregation may overwrite useful object evidence.
Applying cross-type attention before feature aggregation (+Pre-cross) improves mAP by 0.22 and NDS by 0.63 over QInit. However, this is still worse than the proposed post-aggregation QMix. After object evidence has been aggregated, post-aggregation cross-type interaction more effectively consolidates complementary evidence.

\begin{table*}[htbp]
\centering

\begin{minipage}[t]{0.60\textwidth}
\centering
\caption{Analysis of different QMix insertion strategies.}
\resizebox{\textwidth}{!}{
\begin{tabular}{l|cc|cc|cc}
\toprule
Variant & Pre-agg. & Post-agg. & \textbf{mAP}$\uparrow$ & \textbf{NDS}$\uparrow$ & mATE$\downarrow$ & mAOE$\downarrow$ \\
\midrule
QInit & / & / & 54.88 & 61.66 & 0.473 & 0.447 \\
+Post-self & / & self & 54.40 & 61.55 & 0.481 & 0.425 \\
+Post-self + cross & / & self+cross & 54.51 & 61.04 & 0.499 & 0.469 \\
+Pre-cross & cross & / & 55.10 & 62.29 & 0.473 & 0.412 \\
\rowcolor[gray]{.92}+Post-cross (QMix) & / & cross & \textbf{55.51} & \textbf{62.65} & 0.457 & 0.405 \\
\bottomrule
\end{tabular}
}
\label{tab:sec4_ablation_qmix}
\end{minipage}
\hfill
\begin{minipage}[t]{0.385\textwidth}
\centering
\caption{Ablation study of QSwap.}
\resizebox{\textwidth}{!}{
\begin{tabular}{l|c|cc|cc}
\toprule
Setting & $K$ & \textbf{mAP}$\uparrow$ & \textbf{NDS}$\uparrow$ & mATE$\downarrow$ & mAOE$\downarrow$ \\
\midrule
QMix & 20 & 55.51 & 62.65 & 0.457 & 0.405 \\
QMix + 4 samples & 24 & 55.35 & 62.54 & 0.472 & 0.400 \\
QSwap (replace) & 20 & 55.79 & 62.56 & 0.463 & 0.422 \\
\midrule
QSwap + PV & 24 & 55.53 & 62.90 & 0.470 & 0.382 \\
QSwap ($r{=}5$m) & 24 & 55.66 & 62.60 & 0.465 & 0.408 \\
\rowcolor[gray]{.92}QSwap & 24 & \textbf{55.89} & \textbf{63.14} & 0.463 & 0.372 \\
\bottomrule
\end{tabular}
}
\label{tab:sec4_ablation_qswap}
\end{minipage}

\end{table*}

\textbf{QSwap Analysis.}
\cref{tab:sec4_ablation_qswap} presents an ablation study of QSwap. Simply increasing the number of base sampling points $K$ from 20 to 24 does not improve performance over QMix, indicating that the gain does not come from using more samples alone. Using the replacement variant of QSwap instead of the appending variant, which keeps the total number of sampling points unchanged, still improves mAP over QMix, showing the benefit of interactive sampling itself. Applying QSwap to PV sampling hurts performance, as projection-based PV sampling is more sensitive to geometric perturbations. Replacing the query-adaptive radius with a fixed 5m radius also degrades performance, indicating that adaptive geometric constraints are important for selecting reliable interaction neighbors.

\begin{figure}[htbp] \centering
\centering
    \includegraphics[width=0.95\textwidth]{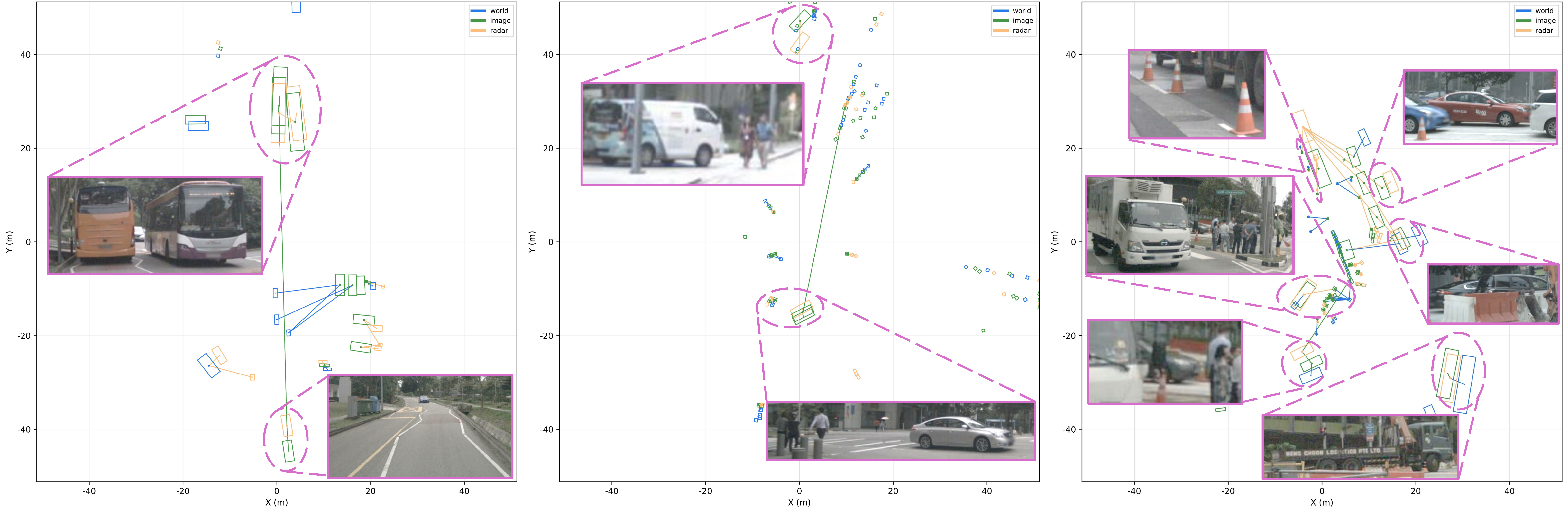}
    \caption{QMix cross-type attention links. High-confidence boxes and their top-2 links are visualized.} \label{fig:sec4_qmix_viz}
\end{figure}

\textbf{QMix Visualization.} We visualize the cross-type attention links in QMix to analyze its behavior (\cref{fig:sec4_qmix_viz}). After filtering queries with confidence above 0.1, we focus on high-confidence queries and show only their top-2 attention links to reveal the most salient associations. Even under this strict filtering, QMix links high-confidence queries with their cross-type counterparts corresponding to the same object, showing that queries with different priors are explicitly associated and fused.

\section{Conclusion} \label{sec:conclusion}

We presented ConFusion, a camera-radar 3D object detector built on a new fusion paradigm, heterogeneous query interaction. ConFusion explicitly controls how heterogeneous queries interact through the proposed QMix and QSwap modules, enabling more effective camera-radar fusion. Experiments on nuScenes demonstrate state-of-the-art performance, while ablation studies and qualitative results support the effectiveness of the proposed design.
Future work may explore improved supervision through assignment strategies or semantic alignment among heterogeneous queries.

\clearpage

{
    \small
    \bibliographystyle{ieeenat_fullname}
    \bibliography{main}
}





\end{document}